# Post-hoc explanation of black-box classifiers using confident itemsets

Milad Moradi[1]

Institute for Artificial Intelligence and Decision Support, Center for Medical Statistics, Informatics, and Intelligent Systems, Medical University of Vienna, Vienna, Austria

`milad.moradivastegani@meduniwien.ac.at`

Matthias Samwald

Institute for Artificial Intelligence and Decision Support, Center for Medical Statistics, Informatics, and Intelligent Systems, Medical University of Vienna, Vienna, Austria

`matthias.samwald@meduniwien.ac.at`

[1] Corresponding author. **Postal address:** Institute for Artificial Intelligence and Decision Support, Währinger Straße 25a, 1090 Vienna, Austria. **Telephone number:** 0043-1-40160-36313

**Abstract**

Black-box Artificial Intelligence (AI) methods, e.g. deep neural networks, have been widely utilized to build predictive models that can extract complex relationships in a dataset and make predictions for new unseen data records. However, it is difficult to trust decisions made by such methods since their inner working and decision logic is hidden from the user. Explainable Artificial Intelligence (XAI) refers to systems that try to explain how a black-box AI model produces its outcomes. Post-hoc XAI methods approximate the behavior of a black-box by extracting relationships between feature values and the predictions. Perturbation-based and decision set methods are among commonly used post-hoc XAI systems. The former explanators rely on random perturbations of data records to build local or global linear models that explain individual predictions or the whole model. The latter explanators use those feature values that appear more frequently to construct a set of decision rules that produces the same outcomes as the target black-box. However, these two classes of XAI methods have some limitations. Random perturbations do not take into account the distribution of feature values in different subspaces, leading to misleading approximations. Decision sets only pay attention to frequent feature values and miss many important correlations between features and class labels that appear less frequently but accurately represent decision boundaries of the model. In this paper, we address the above challenges by proposing an explanation method named Confident Itemsets Explanation (CIE). We introduce confident itemsets, a set of feature values that are highly correlated to a specific class label. CIE utilizes confident itemsets to discretize the whole decision space of a model to smaller subspaces. Extracting important correlations between the features and the outcomes of the classifier in different subspaces, CIE produces instance-wise and class-wise explanations that accurately approximate the behavior of the target black-box. Conducting a set of experiments on various black-box classifiers, and different tabular and textual data classification tasks, we show that our CIE method performs better than the previous perturbation-based and rule-based explanators in terms of the descriptive accuracy (an improvement of 9.3%) and interpretability (an improvement of 8.8%) of the explanations. Subjective evaluations demonstrate that the users find the explanations of CIE more understandable and interpretable than those of the other comparison methods.

# 1. Introduction

In the recent decades, Artificial Intelligence (AI) systems have been widely utilized in many domains such as medicine, military, social media, education, industry, transportation, trading, and smart devices. Machine Learning (ML) refers to a broad range of AI methods that try to produce predictions or estimations for new, unseen data according to what has been learned from previous data (Jordan & Mitchell, 2015). In many applications where the underlying data relationships are complex or nonlinear, black-box and intricate ML models, e.g. deep neural networks, support vector machines, or random forests, obtain higher levels of predictive accuracy than simple models. Deep neural models even perform more accurately than humans in some tasks such as text understanding and image analysis (Zhang, Yang, Chen, & Li, 2018). In spite of their high performance, the applicability of black-box ML models is limited by the inability to understand their inner workings or to explain why certain predictions are made (Adadi & Berrada, 2018; Montavon, Lapuschkin, Binder, Samek, & Müller, 2017). In order to tackle this problem, eXplainable AI (XAI) methods have been developed. XAI methods allow users to understand why a system makes certain predictions, why it fails, what types of errors the system is prone to, and what biases exist in the model or data (Guidotti, Monreale, Ruggieri, Turini, et al., 2018; Murdoch, Singh, Kumbier, Abbasi-Asl, & Yu, 2019).

XAI methods can be divided into two broad categories of model-based and post-hoc methods (Murdoch et al., 2019). Model-based explainability, or explainability by design, refers to designing simple and transparent AI models whose inner working and decision logic can be easily represented and interpreted. Decision trees, linear regression models, decision lists, and fuzzy inference systems are typical AI models that are explainable by design (Holzinger, Biemann, Pattichis, & Kell, 2017). This type of methods are useful when the underlying data relationships are not complex, hence the simple models can fit the data well. On the other hand, intricate black-box models should be designed and implemented when the data convey higher degrees of complexity or nonlinearity. In this case, post-hoc explainability is employed to extract information about what relationships the model has learned (Guidotti, Monreale, Ruggieri, Turini, et al., 2018).

A post-hoc XAI method receives a trained and/or tested AI model as input, then generates useful approximations of the model's inner working and decision logic by producing understandable representations in the form of feature importance scores, rule sets, heatmaps, or natural language. Many post-hoc methods try to disclose relationships between feature values and outputs of a prediction model, regardless of its internals (Guidotti, Monreale, Ruggieri, Pedreschi, et al., 2018; Guidotti, Monreale, Ruggieri, Turini, et al., 2018; Ribeiro, Singh, & Guestrin, 2016). This helps users identify the most important features in a ML task, quantify the importance of features, reproduce decisions made by the black-box model, and identify biases in the model or data.

Some post-hoc methods, such as Local Interpretable Model-agnostic Explanations (LIME) (Ribeiro et al., 2016), extract feature importance scores by perturbing real samples, observing the change in the ML model's output given the perturbed instances, and building a local simple model that approximates the original model's behavior in the neighborhood of the original samples. One drawback of such methods is that the neighboring instances are produced by randomly perturbing feature values, without taking into account the local distribution of feature values and density of class labels in the neighborhood (Guidotti, Monreale, Ruggieri, Pedreschi, et al., 2018; Ribeiro, Singh, & Guestrin, 2018). In fact, behavior of the model is approximated with respect to randomly generated feature values that may not appear in real samples.

Other class of explanators rely on extracting decision sets or decision lists that represent the black-box model's decision logic in the form of if-then rules (Lakkaraju, Bach, & Leskovec, 2016; Lakkaraju, Kamar, Caruana, & Leskovec, 2017, 2019). Since the number of possible decision rules may be very large for many datasets, the problem of extracting an optimal subset of rules is often modeled as an optimization problem with two main objectives, i.e. the classification accuracy and the overall interpretability (Lakkaraju et al., 2016). However, this approach may impose some limitations. First, the optimization procedure may cause the instance- or class-wise interpretability to be sacrificed in favor of the global interpretability. Second, no measure of importance or confidence is provided, while it may be highly important to know the importance of features or the level of confidence when a feature value is deemed to be associated with a class label. Third, frequent itemsets that represent the most frequently appearing feature values within the whole input space are used as the antecedents and consequents of decision rules. This leads to many feature values to be disregarded because they do not frequently appear in the whole input space, whereas they are highly discriminative in one or few subspaces. Fourth, the overall interpretability objective used by the optimization procedure tries to filter out overlapping decision sets in order to minimize ambiguity. However, a main goal of explainability is revealing uncertain and ambiguous decision boundaries in the model or data. Knowing the uncertain parts and the range of possible behaviors of a model, it becomes easier to constrain the model's parameters or restrict the model's behavior towards what is known within a domain (Tim Menzies, Chiang, Feather, Hu, & Kiper, 2003).

In order to address the above challenges, we propose an explanation method named Confident Itemsets Explanation (CIE) that utilizes confident itemsets to represent instance- and class-wise decision boundaries of a black-box classifier. Experimental results show that our confident itemset-based explanation method can perform better than the other explanators in terms of descriptive accuracy and interpretability of local, post-hoc explanations. The confident itemsets can quantify the strength of local relationships between individual or a set of features and class labels assigned by the target classifier. The method is also able to represent other possible predictions for a data record with respect to the black-box model's decision logic in overlapping subspaces of the input space. In this way, the user can be provided with information about the level of uncertainty and ambiguity in the model or dataset. The CIE method is model-agnostic, i.e. it does not depend on the underlying black-box model.

## 2. Related work

Predictive accuracy metrics, e.g. precision and recall, may not be reliable enough to assess the usefulness of a ML model (Miotto, Wang, Wang, Jiang, & Dudley, 2018). For many tasks, in order to trust a ML model and use it for making real-world decisions, it is needed to understand what relationships the model has learned, how the model produces its outcomes, how the model's decision logic differs in different parts of the feature space, possible biases in the data and model, and the collective influence of features on the model's output. This information can be revealed by a XAI method through exploring the inner working of the underlying black-box model or searching for findings in the data that provide information about the black-box model's decisions (Guidotti, Monreale, Ruggieri, Turini, et al., 2018). XAI methods may vary based on several properties, e.g. the explainability problem at hand, the underlying black-box model, the underlying data modality, and the type of explanator that is used to open the black-box (Du, Liu, & Hu, 2019; Guidotti, Monreale, Ruggieri, Turini, et al., 2018).

Model-based explanation refers to designing transparent ML models that are explainable on their own (Murdoch et al., 2019). A main problem with transparent models is that they are not able to fit complex and non-linear relationships in data, leading to a reduction in predictive accuracy. However, different methods have tried to come up with a tradeoff between model complexity and explainability. The Bayesian Rule Lists (BRL) method (Letham, Rudin, McCormick, & Madigan, 2015) identifies different partitions of a feature space and defines decision logics within each partition using if-then rules. If-then rules can be also represented in the form of short and non-overlapping decision sets that discretize an input space and define separate decision logics within every subspace (Lakkaraju et al., 2016). A prototype-based method (Kim, Rudin, & Shah, 2014) generates a discrete mixture model representing the underlying structure of instances in the form of a set of clusters. Each cluster is characterized by a prototype and a subspace feature indicator, every instance is represented as a mixture of different prototypes, and important features of related prototypes are collected to generate an explanation for every subspace in the dataset. A general approach to building a globally transparent model is to combine several transparent local classifiers and construct a complex model that covers the whole decision space of the given classification task (Parvin, Alinejad-Rokny, Minaei-Bidgoli, & Parvin, 2013; Parvin, MirnabiBaboli, & Alinejad-Rokny, 2015).

Post-hoc explanation, also known as the reverse engineering approach, tries to reconstruct explanations for decisions made by a black-box (Guidotti, Monreale, Ruggieri, Turini, et al., 2018).

Post-hoc explainability can be further divided into global and local methods. Global explanations concern understanding the overall logic and behavior of a black-box model, while local explanations try to find correlations between feature values of a record and an outcome. Global methods may use different types of explanators to open various black-boxes for different problems, e.g. decision trees for explaining random forests (Zhou & Hooker, 2016), prototypes to explain ensemble models (Tan, Hooker, & Wells, 2016), feature importance scores (Vidovic, Görnitz, Müller, & Kloft, 2016), decision trees for identifying training samples responsible for mispredictions (Krishnan & Wu, 2017), and minimal feature adjustment for reverting a class label (Tolomei, Silvestri, Haines, & Lalmas, 2017).

Some local explanation methods try to understand the inner working of deep neural networks using layer-wise relevance propagation (Bach et al., 2015), injecting noise into the input (Fong & Vedaldi, 2017), decomposing the function learnt by a black-box into simpler sub-functions (Montavon et al., 2017), a difference-to-reference approach to feature importance estimation (Shrikumar, Greenside, & Kundaje, 2017), and saliency heatmaps as feature importance visualization (Zintgraf, Cohen, Adel, & Welling, 2017).

One approach to producing local explanations is to analyze the input-output behavior of a black-box model regardless of its internals. LIME (Ribeiro et al., 2016) generates dummy records by perturbing an instance, then approximates the local behavior of the original model in the vicinity of the perturbed instance. The problem is that LIME does not take into account the distribution of feature values. In fact, it approximates a model based on randomly perturbed feature values that may never appear in a record in the dataset. In contrast to LIME, our method approximates correlations between features and an outcome based on real values appeared in the dataset. Our CIE method only considers real samples and predictions made by a black-box classifier. This tackles the problem of black-box approximations based on unrealistic perturbations. Similar to LIME, our CIE explanation method is model-agnostic; it can produce explanations for various black-box models.

Itemset and association mining has been effectively used to discover local patterns in big datasets (Larose & Larose, 2014). Frequent itemsets were already used to discretize the input space of a classification problem (Letham et al., 2015). However, a challenge with previous itemset-based explanation methods, e.g. MUSE (Lakkaraju et al., 2019), is that frequent itemsets only reflect those input-output patterns that frequently appear in a dataset. They disregard infrequent patterns that accurately represent correlations in a small subspace. This issue is more problematic when dealing with complex data relationships, more specifically with text data since frequent words usually have no discriminate power in most classification tasks. On the other hand, our method relies on confident itemsets, i.e. itemsets that accurately represent the local behavior of a model in different parts of the input space. This strategy enables our explanation method to deal with difficulties of approximating more complex correlations, such as multi-class text datasets.

The confidence was already shown to be more effective than the support for pruning domain variables in an association rule learner (Tim Menzies & Hu, 2006; T. Menzies & Ying, 2003). In contrast to frequent itemsets that only consider highly frequent feature values to extract associations between features and classes, confident itemsets precisely approximate relationships between features and class labels. The feature space is discretized into small subspaces, such that in every subspace confident itemsets specify decision boundaries of a class. Confident itemsets can also reveal relationships between multiple feature values and a target class label. In comparison to the decision set methods that use frequent itemsets and optimize the accuracy against the overall interpretability, our method can produce concise and easily understandable explanations without decreasing the descriptive accuracy. Similar to Anchors (Ribeiro et al., 2018), our CIE explanator can produce concise explanations by extracting minimal sets of feature values that precisely approximate the local behavior of a black-box.

## 3. Confident itemsets explanation method

For many tasks and datasets, decision boundaries of a classification model may be too complicated such that the whole model cannot be explained by a concise and interpretable representation. In this case, a solution is to discretize the feature space and construct explanations for smaller subspaces, then the explanations can be combined to represent the black-box model's behavior in larger subspaces. Our explanation method works based on the above idea. We first describe how confident itemsets are constructed to reveal relationships between feature values and class labels for individual instances. Then we show how class-wise explanations are built upon instances-wise explanations.

### 3.1. Extracting confident itemsets

Let $f : X \rightarrow C$ be a black-box classifier, $X_i \in X$ be an instance consisting of $N$ tuples, such that $X_i = (f_1, v_1), \ldots, (f_N, v_N)$, where $f_n$ is $n_{th}$ feature and $v_n$ is the corresponding value, and $C=\{C_1, C_2, \ldots, C_Q\}$ be the set of classes of the classification problem. Given a class label $Y_m \in C$ predicted by $f$ as the prediction for $X_m$, the goal is to explain the local behavior of $f$ using a set of itemsets that refer to those feature-value pairs $(f_n, v_n)$ that are highly correlated with class $C_q$ that corresponds to $Y_m$. Given a dataset $D=\{X_1, X_2, \ldots, X_M\}$ containing $M$ instances, and $Y=\{Y_1, \ldots, Y_M\}$ containing $M$ class labels such that $Y_m$ is the class label predicted for $X_m$ by the black-box classifier $f$, the explanation method begins by discretizing the feature space $S$ into subspaces $\{S_1, \ldots, S_Q\}$ such that every subspace $S_q$ contains those instances that were classified in class $C_q$. Confident itemsets, i.e. those feature values that are highly correlated with a certain class, are discovered within each subspace and are represented in the form of *<feature, operator, value>* triples, e.g. *<age, <=, 30>* or *<Education, =, High-school>*.

Hereafter, we explain the confident itemset mining procedure for a tabular dataset, i.e. a mixture of categorical and numerical features and their corresponding values for a set of records. However, the same procedure can also apply to a textual dataset. When working with a tabular dataset, every data record is represented as a fixed number of triples *<feature₁, operator₁, value₁>, …, <featureₙ, operatorₙ, valueₙ>*, where $N$ is the number of features and every triple is considered as an item. When working with a text dataset, every text instance is represented as a set of words *<Word₁>, …, <Word_G>*, where $G$ can vary for every instance. Every word is considered as an individual item. Figure 1 shows itemset-based representations for tabular and text data instances.

A tabular data record from the Adult dataset and the respective itemset-based representation is shown in Figure 1(a). Every record in this dataset contains 14 numerical and categorical features representing different information of a person, e.g. age, education, marital status, occupation, capital gain, and a binary class label that specifies whether the person has an annual income more than \$50K. A categorical feature value such as *workclass=Private* is simply represented as an item *<workclass, =, Private>*. Numerical features are divided into ranges based on the input given by the user. For example, given an input dividing the feature *hours-per-week* into ranges *<=15, [16, 30], [31, 45], [46, 60], [61, 75]*, and *>=76*, a numerical feature value such as *hours-per-week=60* is represented as *<hours-per-week, in, [46, 60]>*. A textual data record from the TREC question classification dataset and the respective itemset-based representation is shown in Figure 1(b). Every instance in this dataset is a text question with a semantic class label referring to the topic of the question. The text is tokenized and every token is represented as an item.

Extracting confident itemsets is done through an iterative algorithm. A set of confident $K$-itemsets is extracted for every class $C_q$ in $K_{\text{th}}$ iteration. An item is a triple *<F, O, V>* representing a feature, an operator, and the corresponding value that appeared in the dataset. Given a class $C_q$, a confident $K$-itemset $ci$ is a set of $K$ distinct items that satisfies two criteria: 1) the confidence property of $ci$ within the class $C_q$ must be equal to or greater than a confidence threshold *min_conf*, and 2) every subset of $ci$ must be a confident itemset within the class $C_q$. The confidence property of $ci$ is computed within class $C_q$ as follows:

$$Confidence\big(ci, C_q\big) = \frac{P(ci|C_q)}{P(ci)} \tag{1}$$

where $P(ci)$ is the probability of observing $ci$ in dataset $D$, and $P(ci|C_q)$ is the probability of observing $ci$ in instances belonging to class $C_q$.

(a)

Tabular record:

| age | workclass | fnlwgt | education | education-num | marital-status | occupation | relationship | race | sex | capital-gain | capital-loss | hours-per-week | native-country |
|---|---|---|---|---|---|---|---|---|---|---|---|---|---|
| 30 | Private | 167476 | Some-college | 10 | Married-civ-spouse | Transport-moving | Husband | White | Male | 0 | 0 | 60 | United-States |

Itemset-based representation:

*<**age**, <=, 30>, <**workclass**, =, Private>, <**fnlwgt**, in, [150001, 250000]>, <**education**, =, Some-college>, <**education-num**, in, [9, 12]>, <**marital-status**, =, Married-civ-spouse>, <**occupation**, =, Transport-moving>, <**relationship**, =, Husband>, <**race**, =, White>, <**sex**, =, Male>, <**capital-gain**, <=, 0>, <**capital-loss**, <=, 0>, <**hours-per-week**, in, [46, 60]>, <**native-country**, =, United-States>*

(b)

Text record:

"What is the fastest commercial automobile that can be bought in the US?"

Itemset-based representation:

<What>, <is>, <the>, <fastest>, <commercial>, <automobile>, <that>, <can>, <be>, <bought>, <in>, <the>, <US>

**Figure 1.** (a) The itemset-based representation of a tabular record from the Adult dataset. Every pair of a feature and its value is considered as an item and represented as a triple *<feature, operator, value>*. Numerical features are divided into ranges based on the input given by the user. (b) The itemset-based representation of a text record from the TREC question classification dataset. Every single word is considered as an item. The datasets are described in Section 3.

First, an empty set $CI_q$ of confident itemsets is initialized for every class $C_q$. Every numerical feature in the dataset is divided into ranges based on the user input, in order to select one operator when representing the feature and its value as a *<feature, operator, value>* triple. The itemset mining algorithm begins by extracting confident 1-itemsets, i.e. confident itemsets containing only one item. In every iteration, extracted confident $K$-itemsets are added to $CI_q$ and $K$ increases by one. The itemset mining algorithm continues until $K$ reaches a predefined value, or it stops when no confident itemset is extracted in the latest iteration. Every extracted confident itemset $ci$ is represented in the form of a set of triples $\{<F_1, O_1, V_1>, <F_2, O_2, V_2>, ..., <F_J, O_J, V_J>\}$ where $F_j$ refers to a feature, $O_j$ is an operator, and $V_j$ refers to the corresponding value. The pseudo-code for the confident itemset mining algorithm is presented by Algorithm 1.

Figure 2 shows the confident itemsets extracted from the predictions of a Long Short-Term Memory (LSTM) text classifier for the class "Entity:substance" in the TREC question classification dataset. Figure 3 shows the confident itemsets extracted from the predictions of a multi-layer perceptron classifier for the class "group: <=50K" in the Adult dataset.

**Algorithm 1.** The confident itemset mining algorithm employed by the CIE explanation method.

---

1: **Input:** dataset $D$, minimum confidence threshold *min_conf*, maximum number of items per itemset *max_K*,

2: **Output:** confident itemsets *CI*

3: $CI=\varnothing$

4: divide every numerical feature in the dataset into ranges based on the user input

5: **for** every class $C_q$ **do**

6:    $CI_q=\varnothing$

7:    **for** every distinct pair $((f_n, v_n) \mid (f_n, v_n) \in X_m$ and $f(X_m)=C_q)$ **do**

8:       create an item $I_n$ and represent it as a triple $<F_n, O_n, V_n>$

9:       compute a Confidence value *Confidence($I_n$, $C_q$)*

10:       **if** *Confidence($I_n$, $C_q$) >= min_conf* **then** $CI_q=CI_q \cup I_n$

11:    **end for**

12:    $K=2$

13:    **while** no confident itemset is added to $CI_q$ or $K==max\_K$

14:       **for** every $(ci_j \in CI_q \mid Size(ci_j)=K\text{-}1)$ **do**

15:          **for** every $(ci_p \in CI_q \mid ci_p \neq ci_j$ and $Size(ci_p)=K\text{-}1)$ **do**

16:             **if** $ci_j$ and $ci_p$ have $K\text{-}2$ items in common **then**

17:                *Candidate_itemset*$=ci_j \cup ci_p$

18:                   **if** *Confidence(Candidate_itemset, $C_q$) >= min_conf* **and** every subset of *Candidate_itemset* satisfies the minimum confidence criterion **then** $CI_q=CI_q \cup$ *Candidate_itemset*

19:          **end for**

20:       **end for**

21:       $K=K+1$

22:    **end while**

23:    $CI=CI \cup CI_q$

24: **end for**

25: **return** *CI*

---

| Class: ENTY:substance | | | | | |
|---|---|---|---|---|---|
| *Itemset* | *Confidence* | *Class_support* | *Itemset* | *Confidence* | *Class_support* |
| <substance> | 1.0 | 0.117 | <solid> | 0.78 | 0.029 |
| <elements> | 0.8 | 0.117 | <liquid> | 0.8 | 0.029 |
| <consist> | 1.0 | 0.058 | <saliva> | 1.0 | 0.029 |
| <hardest> | 0.93 | 0.058 | <charcoal> | 1.0 | 0.029 |
| <crust> | 1.0 | 0.029 | <hardest>, <substance> | 1.0 | 0.058 |
| <ribavirin> | 1.0 | 0.029 | <ribavirin>, <consist> | 1.0 | 0.029 |
| <organic> | 0.87 | 0.029 | <solid>, <liquid> | 1.0 | 0.029 |
| <sulfur> | 1.0 | 0.029 | <elements>, <crust> | 1.0 | 0.029 |
| <seawater> | 1.0 | 0.029 | <elements>, <seawater> | 1.0 | 0.029 |
| <magnesium> | 1.0 | 0.029 | <charcoal>, <sulfur> | 1.0 | 0.029 |

**Figure 2.** Confident itemsets extracted from the predictions of a LSTM text classifier for the class "Entity:substance" in the TREC question classification dataset. The minimum confidence threshold is set to 0.7 in this example. Fourteen 1-itemsets and six 2-itemsets are represented in this example. The *confidence* measure quantifies the strength of the relationship between an itemset and the respective class label. The *class_support* measure quantifies how frequently the itemset appears in the respective subspace.

| Class: <=50K | | | | | |
|---|---|---|---|---|---|
| *Itemset* | *Confidence* | *Class_support* | *Itemset* | *Confidence* | *Class_support* |
| <age, in, [21, 30]> | 0.918 | 0.303 | <hours-per-week, in, [31, 45]> | 0.776 | 0.629 |
| <occupation, =, Other-service> | 0.958 | 0.128 | <education-num, in, [9, 12]>, <education, =, Some-college> | 0.817 | 0.243 |
| <sex, =, Female> | 0.891 | 0.386 | <hours-per-week, in, [31, 45]>, <education, =, Some-college> | 0.816 | 0.142 |
| <capital-gain, <=, 0> | 0.958 | 0.794 | <workclass, =, Private>, <sex, =, Female> | 0.907 | 0.285 |
| <capital-loss, <=, 0> | 0.970 | 0.774 | <capital-gain, <=, 0>, <occupation, =, Other-service> | 0.964 | 0.124 |
| <education-num, in, [9, 12]> | 0.818 | 0.672 | <capital-loss, <=, 0>, <age, in, [21, 30]> | 0.923 | 0.295 |
| <education, =, Some-college> | 0.817 | 0.243 | <capital-loss, <=, 0>, <workclass, =, Private>, <sex, =, Female> | 0.913 | 0.278 |
| <workclass, =, Private> | 0.783 | 0.715 | <capital-loss, <=, 0>, <education-num, in, [9, 12]>, <education, =, Some-college> | 0.826 | 0.236 |

**Figure 3.** Confident itemsets extracted from the predictions of a multi-layer perceptron classifier for the class "group: <=50K" in the Adult dataset. The minimum confidence threshold is set to 0.7 in this example. Nine 1-itemsets, five 2-itemsets, and two 3-itemsets are represented in this example. The *confidence* measure quantifies the strength of the relationship between an itemset and the respective class label. The *class_support* measure quantifies how frequently the itemset appears in the respective subspace.

## 3.2. Instance-wise explanations

Confident itemsets extracted by the itemset mining algorithm can be easily used to approximate the black-box model's behavior for individual predictions. The itemsets can show how feature values are related to class labels in different subspaces. They can also show how different features are related together in a subspace. Every confident itemset $ci$ contains a set of triples $\{ <F_1, O_1, V_1>, <F_2, O_2, V_2>, ..., <F_J, O_J, V_J> \}$ and is associated with a confidence value within a class $C_q$. A support property is defined for every itemset to show how frequent the itemset is in different subspaces, also in the whole feature space. Given a confident itemset $ci$, an overall support property is defined as the probability of observing $ci$ within the whole feature space, as follows:

$$Overall\_support(ci) = \frac{Count(ci)}{M} \qquad (2)$$

where $Count(ci)$ counts the number of instances that contain confident itemset $ci$, and $M$ is the total number of instances in dataset $D$. A support value is also computed for confident itemset $ci$ in class $C_q$, as follows:

$$Class\_support(ci, C_q) = \frac{Count(ci, C_q)}{M_q} \qquad (3)$$

where $Count(ci, C_q)$ counts the number of instances that contain confident itemset $ci$ and are assigned to class $C_q$, and $M_q$ is the total number of instances assigned to class $C_q$. The overall and class-wise support values are used to assess the importance of confident itemsets and rank them based on their frequency when producing class-wise and global explanations.

Given a black-box classifier $f$, a dataset $D=\{X_1, X_2, ..., X_M\}$ containing $M$ instances, a set of classes $C=\{C_1, C_2, ..., C_q\}$ defined by the classification problem, a set of class labels $Y=\{Y_1, Y_2, ..., Y_M\}$ assigned by the black-box classifier such that $f(X_m)=Y_m$ and $Y_m \in C$, a set of confident itemsets $CI=\{CI_1, CI_2, ..., CI_Q\}$ such that $CI_q$ holds a set of confident itemsets $\{ci_1, ci_2, ..., ci_J\}$ extracted for class $C_q$, an explanation $E_m =\{<ci_1, ci_2, ..., ci_U>, Y'_m\}$ is produced for instance $X_m$ such that $<ci_1, ci_2, ..., ci_U>$ is a set of confident itemsets and $Y'_m$ is the class label assigned to instance $X_m$ by the explanation model. The set $<ci_1, ci_2, ..., ci_U>$ is constructed through searching every $CI_q \in CI$ and extracting those itemsets that occur in $X_m$. A confidence score is produced for $E_m$ within every class $C_q$ that has at least one itemset in the set $<ci_1, ci_2, ..., ci_U>$, as follows:

$$Confidence\_score(E_m, C_q) = \sum_{u=1}^{U} Confidence\,(ci_u, C_q) \qquad (4)$$

where $Confidence\_score(E_m, C_q)$ is the confidence score of explanation $E_m$ in class $C_q$, and $Confidence(ci_u, C_q)$ is the confidence value of itemset $ci_u$ in class $C_q$.

Finally, the explanation model selects a class label for instance $X_m$ by assigning the class with the highest confidence score to $Y'_m$. When a class $C_q$ obtains the highest confidence score, this means some feature values appearing in instance $X_m$ are highly correlated with the class label in the subspace characterized by $C_q$.

Figure 4 and Figure 5 show instance-wise explanations generated by the CIE method for predictions of a LSTM classifier on the TREC question classification dataset and a multi-layer perceptron classifier on the Adult dataset, respectively. In each figure, one correctly predicted and one mispredicted record by the respective black-box classifier are represented.

## 3.3. Class-wise explanations

An approach to producing an explanation for a single class is aggregating instance-wise explanations that belong to the class. However, for some tasks or datasets, extremely large explanations may be produced by aggregating every single instance-wise explanation, leading to low degrees of interpretability. A possible solution is selecting an optimal subset of confident itemsets that accurately approximate the black-box model's behavior in the respective subspace using an interpretable representation. To this end, fidelity, interpretability, and coverage measures must be quantified for subsets of itemsets.

(a)

Text record:

| "Where is Mile High Stadium?" |
|---|

Prediction:

| Real class | Predicted by black-box | Approximated by CIE |
|---|---|---|
| LOC:other | LOC:other | LOC:other |

Explanation:

| Class: LOC:other | | Class: NUM:count | |
|---|---|---|---|
| Score: 2.555 | | Score: 0.666 | |
| Itemset | Confidence | Itemset | Confidence |
| <where> | 0.888 | <mile> | 0.666 |
| <stadium> | 0.666 | | |
| <where>, <stadium> | 1.0 | | |

(b)

Text record:

| "What singer 's theme song was When the Moon Comes over the Mountain?" |
|---|

Prediction:

| Real class | Predicted by black-box | Approximated by CIE |
|---|---|---|
| HUM:ind | LOC:mount | LOC:mount |

Explanation:

| Class: LOC:mount | | Class: NUM:date | |
|---|---|---|---|
| Score: 0.666 | | Score: 0.661 | |
| Itemset | Confidence | Itemset | Confidence |
| <mountain> | 0.666 | <when> | 0.661 |
| | | | |
| | | | |

**Figure 4.** The instance-wise explanations approximated by the CIE method for two text records from the TREC question classification dataset, (a) correctly predicted and (b) mispredicted by a black-box classifier. The minimum confidence threshold is set to 0.6 in this example. In every explanation, the left column shows the class with the highest confidence score selected by CIE as the approximated class label. The right column shows an alternative class label that achieved lower confidence score than the selected class. The itemsets that provide evidence in favor of a particular class label are listed below the corresponding column.

**Figure 5.** The instance-wise explanations approximated by the CIE method for two tabular records from the Adult dataset, (a) correctly predicted and (b) mispredicted by a black-box classifier. The minimum confidence threshold is set to 0.5 in this example. In the explanation (b), the left column shows the class with the highest confidence score selected by CIE as the approximated class label. The right column shows an alternative class label that achieved lower confidence score than the selected class. The itemsets that provide evidence in favor of a particular class label are listed below the corresponding column.

Given a class $C_q$, a set of confident itemsets $CI_q=\{ci_1,\ ci_2,\ ...,\ ci_J\}$ extracted for $C_q$, a class-wise explanation $CE_q=\{ci_1,\ ci_2,\ ...,\ ci_B\}$ is defined as an optimal subset of $CI_q$ that optimizes fidelity, interpretability, and coverage measures. The fidelity metric quantifies how accurate class-wise explanation $CE_q$ can mimic black-box model $f$ in terms of assigning class labels to instances, as follows:

$$Fidelity(CE_q) = \frac{\sum_{m=1}^{M} X_m \in D \mid f(X_m) = C_q \ and \ f(X_m) = E(X_m)}{M_q} \tag{5}$$

where $E(X_m)$ is the explanation model's prediction for instance $X_m$, and $M_q$ is the number of instances in dataset $D$ predicted by black-box model $f$ as class $C_q$.

Four measures, i.e. *Size*, *NumItems*, *MaxLength*, and *ItemsetOverlap*, are defined in order to quantify the interpretability of class-wise explanations. The size of explanation $CE_q$ is computed as the number of confident itemsets that belong to $CE_q$, as follows:

$$Size(CE_q) = \sum_{b=1}^{B} ci_b \tag{6}$$

The number of items in $CE_q$ is computed by summing up the size of every confident itemset that belong to $CE_q$, as follows:

$$NumItems(CE_q) = \sum_{b=1}^{B} Size(ci_b) \qquad (7)$$

where *Size(ci_b)* is the number of items in confident itemset *ci_b*.

The maximum length of *CE_q* is defined as the maximum size of a confident itemset in *CE_q*, as follows:

$$MaxLength(CE_q) = Max_{b=1}^{B} Size(ci_b) \qquad (8)$$

An itemset overlap measure is defined as the number of confident itemset pairs that have at least one item in common, as follows:

$$ItemsetOverlap(CE_q) = \sum_{b=1}^{B} \sum_{h=1}^{H} ci_b, ci_h \mid Overlap(ci_b, ci_h) = True \text{ and } ci_h \in CE_q \qquad (9)$$

where *Overlap(ci_b, ci_h)* is a binary function that returns *True* when confident itemsets *ci_b* and *ci_h* have at least one item in common.

A coverage measure is also defined for class-wise explanation *CE_q* as the number of instances predicted as class *C_q* that are covered by at least one confident itemset belonging to *CE_q*, as follows:

$$Coverage(CE_q) = \sum_{m=1}^{M} X_m \in D \mid f(X_m) = C_q \text{ and } Cover(CE_q, X_m) = True \qquad (10)$$

where *Cover(CE_q, X_m)* is a binary function that returns *True* when at least one confident itemset *ci_b* ∈ *CE_q* occurs in instance *X_m*.

Next, a non-negative reward function is defined for every measure. For those measures that lower values are preferred, the computed value is subtracted from its upper bound value. The reward functions are defined as follows:

$$f_1(CE_q) = Fidelity(CE_q) \qquad (11)$$

$$f_2(CE_q) = Max\left(Size(CE_q)\right) - Size(CE_q) \qquad (12)$$

$$f_3(CE_q) = Max\left(NumItems(CE_q)\right) - NumItems(CE_q) \qquad (13)$$

$$f_4(CE_q) = Max\left(MaxLength(CE_q)\right) - MaxLength(CE_q) \qquad (14)$$

$$f_5(CE_q) = Max\left(ItemsetOverlap(CE_q)\right) - ItemsetOverlap(CE_q) \qquad (15)$$

$$f_6(CE_q) = Coverage(CE_q) \qquad (16)$$

Finally, an objective function is formulated using the six reward functions in order to jointly optimize the fidelity, interpretability, and coverage measures. The optimization problem is formulated using the following objective:

$$max_{CE_q \subseteq CI_q} \sum_{i=1}^{6} w_i f_i(CE_q) \tag{17}$$

$$Constraints: Size\big(CE_q\big) \leq \theta_1, NumItems\big(CE_q\big) \leq \theta_2, MaxLength\big(CE_q\big) \leq \theta_3$$

where $w_1$, $w_2$, ..., $w_6$ are non-negative weights that control the relative importance of reward functions. The weights are selected through a cross-validation procedure. The values of $\theta_1$, $\theta_2$, and $\theta_3$ depend on the explainability problem at hand, and should be specified by the end user.

The objective given by Eq. (17) is submodular, non-monotone, non-negative, and non-normal, as proven by Lee et al. (Lee, Mirrokni, Nagarajan, & Sviridenko, 2009) and Lakkaraju et al. (Lakkaraju et al., 2019). Consequently, we utilize the optimization method proposed by Lee et al. (Lee et al., 2009), which relies on approximate local search and guarantees an optimal solution for this type of problem. Algorithm 2 presents a pseudo-code of the optimization procedure used by our CIE explanation method. Finally, the class-wise explanations are used to approximate the behavior of the black-box model in larger subspaces.

**Algorithm 2.** The optimization algorithm employed by the CIE method to produce class-wise explanations.

---

1: **Input:** objective $f$, set of confident itemsets $CI_q$, parameter $\delta$, number of constraints $k$

2: **Output:** class-wise explanation $CE_q$

3: $E_1 = CI_q$

4: **for** $i \in \{1, 2, ..., k+1\}$ **do**

5:     $X = E_i$, $n = |X|$, $S_i = \varnothing$

6:     $S_i \leftarrow$ the element with the maximum value for objective $f$

7:     **while** a delete or update operation increases the value of $S_i$ by a factor of at least $(1 + \frac{\delta}{n^4})$ **do**

8:         (delete operation) **if** there is an element $a \in S_i$ such that $f(S_i \backslash a) \geq (1 + \frac{\delta}{n^4})f(S_i)$ **then**

9:             $S_i \leftarrow S_i \backslash a$

10:         (update operation) **if** there is an element $b \in X \backslash S_i$ and an element $a_j \in S_i$ such that

11:         $(S_i \backslash a_j) \cup \{b\}$ (for $1 \leq j \leq k$) satisfies all the $k$ constrains and

12:         $f(S_i \backslash \{a_1, a_2, ..., a_k\} \cup \{b\}) \geq (1 + \frac{\delta}{n^4})f(S_i)$ **then**

13:             $S_i \leftarrow S_i \backslash \{a_1, a_2, ..., a_k\} \cup \{b\}$

14:     **end while**

15:     $E_{i+1} = E_i \backslash S_i$

16: **end for**

17: $CE_q \leftarrow \max\{f(S_1), f(S_2), ..., f(S_{k+1})\}$

18: **return** $CE_q$

---

# 4. Experimental results and discussion

We used three tabular datasets. The **Adult** dataset[2] from UCI Machine Learning Repository contains more than 48,000 records with demographic information used to classify persons into two classes indicating whether every individual makes an annual income more than $50K or not. The **Compas** dataset[3] from ProPublica comprises more than 37,000 records of defendants and their recidivism risk scores. Every instance is classified as a low, medium, or high risk individual. The **Thyroid Disease** (TD) dataset[4] from UCI Machine Learning Repository includes information of more than 2,000 individuals and their biomarkers that are classified into two classes, i.e. sick and negative.

We also used three text classification datasets. The **TREC Question Classification** (TREC-QC) dataset[5] contains more than 6,000 questions classified into 50 classes specifying the types of questions. The **Stanford Sentiment Treebank** (SST) dataset (Socher et al., 2013) includes more than 200,000 phrases extracted from 'Rotten Tomatoes' film reviews. The phrases are classified into five sentiment classes, i.e. very negative, negative, neutral, positive, and very positive. The **SMS Spam Collection** (SMS-SC) dataset[6] from UCI Machine Learning Repository contains more than 5,000 SMS messages that are classified into spam and non-spam classes.

For the tabular data classification tasks, we used the scikit-learn library (Pedregosa et al., 2011) with default parameters, unless it is stated, to implement three models:

- **Multi-Layer Perceptrons** (MLP) with 'lbfgs' solver, five layers of neurons, and the cross-entropy loss function. A MLP (Murtagh, 1991) is a neural network composed of an input, output, and multiple hidden layers. In every hidden layer, neurons receive inputs from the previous layer, compute a nonlinear transformation of the inputs, and deliver the result to neurons in the next layer. Finally, nodes in the output layer estimate the probability of different class labels based on the computations done by the hidden layers' nodes.

- **Support Vector Machines** (SVM) with RBF kernel. SVMs (Schlkopf, Smola, & Bach, 2018) try to find a hyperplane separating two classes such that the distance between the classes' margin and the hyperplane is maximized. When the classes are not linearly separable, SVM kernels use mathematical transformations to add more dimensions to the feature space. A hyperplane in the higher-dimensional space finds a linear separation between the classes.

- **Gradient Boosted Decision Trees** (GBDT) with 100 weak learners. GBDT (Hastie, Tibshirani, & Friedman, 2009) is a type of boosting methods that gradually creates a powerful classification model as an ensemble of weak prediction models. In a step-wise procedure, the model is built by selecting the best combination of a new model and previous models such that the overall prediction error is minimized.

---

[2] https://archive.ics.uci.edu/ml/datasets/Adult
[3] https://www.propublica.org/datastore/dataset/compas-recidivism-risk-score-data-and-analysis
[4] https://archive.ics.uci.edu/ml/datasets/Thyroid+Disease
[5] https://cogcomp.seas.upenn.edu/Data/QA/QC/
[6] https://archive.ics.uci.edu/ml/datasets/sms+spam+collection

We also implemented Long Short-Term Memory (LSTM) (Hochreiter & Schmidhuber, 1997) text classifiers for the text classification tasks. Word embeddings (Mikolov, Sutskever, Chen, Corrado, & Dean, 2013) were used to represent the input instances as continuous vectors, but our explanation method uses words to explain the predictions. Each dataset was split into training and test sets; the black-box models were trained on training data and explanations were produced for predictions made on test data.

We compare our explanation method with Local Interpretable Model-agnostic Explanations (LIME) (Ribeiro et al., 2016) that is publicly available. We also implemented Model Understanding through Subspace Explanations (MUSE) (Lakkaraju et al., 2019) with respect to the method described in the respective paper. We also compared our CIE method against greedy and random approaches. The greedy method, which is similar to the heuristic proposed by Martens et al. (Martens & Provost, 2014), chooses the most *K* important features within every class. The random method selects *K* random features within every class as explanations for instances belonging to that class.

The source code of the CIE explanation method is available at: https://github.com/mmoradi-iut/CIE-explanation-method.

## 4.1. Fidelity to the black-box's predictions

Fidelity, also known as descriptive accuracy (Murdoch et al., 2019), measures how accurate an explanation method can mimic the behavior of a black-box classifier in terms of assigning class labels to data records. Fidelity has been widely used as a measure to evaluate XAI systems (Guidotti, Monreale, Ruggieri, Pedreschi, et al., 2018; Lakkaraju et al., 2017; Ribeiro et al., 2016), since a high fidelity score ensures that the explanations can accurately reflect the black-box's decisions, and a low fidelity leads to wrong explanations that are quite useless no matter how understandable and comprehensive the explanations are.

We did no modification in LIME for instance-wise explanations. The maximum number of feature importance scores per instance generated by LIME was set to 10. Instance-wise explanations were produced for MUSE by extracting those decision rules that satisfy the data record being explained. In order to utilize MUSE in text classification experiments, we implemented a version of MUSE that considers presence of words as predicates in decision rules, instead of considering feature values. Table 1 presents the fidelity scores obtained by our CIE explanation method and the other comparison methods for instance-wise explanation experiments on the three tabular datasets. Table 2 presents the fidelity results on the three text classification datasets.

**Table 1.** The fidelity scores obtained by the explanation methods for instance-wise explanation experiments on the tabular datasets. The best score obtained on each dataset is shown in bold type.

| | Black-box | Tabular datasets | | |
| --- | --- | --- | --- | --- |
| | | *Adult* | *Compas* | *TD* |
| MUSE | MLP | 0.882 | 0.865 | 0.829 |
| | SVM | 0.861 | 0.874 | 0.811 |
| | GBDT | 0.879 | 0.860 | 0.817 |
| LIME | MLP | 0.840 | 0.822 | **0.896** |
| | SVM | 0.849 | 0.827 | 0.881 |
| | GBDT | 0.853 | 0.810 | 0.873 |
| Greedy | MLP | 0.767 | 0.759 | 0.740 |
| | SVM | 0.762 | 0.754 | 0.735 |
| | GBDT | 0.763 | 0.748 | 0.721 |
| Random | MLP | 0.561 | 0.530 | 0.497 |
| | SVM | 0.511 | 0.545 | 0.499 |
| | GBDT | 0.509 | 0.521 | 0.474 |
| CIE | MLP | **0.917** | **0.904** | 0.882 |
| | SVM | 0.902 | 0.898 | 0.889 |
| | GBDT | 0.894 | 0.901 | 0.870 |

As the results show, our CIE method obtained the highest fidelity scores on two of the tabular and on all the three text classification datasets. As mentioned earlier, LIME uses random perturbations to generate dummy instances in the neighborhood of an instance and approximates a linear model, regardless of the distribution of feature values in the dataset. This may negatively affect the accuracy of local explanations, since they are produced based on dummy instances that do not truly represent feature values that appear in real instances. However, LIME may perform better on datasets with more continuous features, as it obtained the highest fidelity scores on the TD dataset. This may be due to its strategy for sampling uniform random perturbations and the efficiency of the linear regression model in approximating proper feature importance scores when dealing with continuous features that most of them have linear correlation with the target variable.

**Table 2.** The fidelity scores obtained by the explanation methods for instance-wise explanation experiments on the text datasets. The best score obtained on each dataset is shown in bold type.

| | Black-box | Text datasets | | |
| --- | --- | --- | --- | --- |
| | | *TREC-QC* | *SST* | *SMS-SC* |
| MUSE | LSTM | 0.754 | 0.722 | 0.731 |
| LIME | LSTM | 0.859 | 0.840 | 0.832 |
| Greedy | LSTM | 0.744 | 0.715 | 0.723 |
| Random | LSTM | 0.513 | 0.507 | 0.489 |
| CIE | LSTM | **0.925** | **0.891** | **0.910** |

MUSE obtained higher scores than LIME on two tabular datasets, but it could not perform well on text datasets. It uses frequent itemsets as predicates to specify subspace descriptors and decision logics within each subspace. In this way, only the most frequent feature values in tabular datasets, or words in text datasets, are used to discriminate between different subspaces, whereas those feature values that appear more frequently may not have strong discriminative power for discretizing subspaces in many datasets, especially when working with words as items in text datasets.

On the other hand, our CIE method tries to extract confident itemsets that properly convey decision boundaries of every class. Confident itemsets represent combinations of feature values that are highly associated with a specific class label. The results show that confident itemsets can be effectively used to identify discriminative feature values within different subspaces in order to accurately approximate the behavior of various black-box classifiers. Confident itemsets are efficiently applicable to text data, since frequent words may not be helpful in terms of being discriminative in most text datasets. In this case, those words that are highly associated with class labels, even if they are not frequent, are extracted and used to mimic decisions made by a black-box.

In addition to instance-wise experiments, we also assessed performance of the methods in producing class-wise explanations. Given a class, a class-wise explanation was produced for LIME by collecting explanations generated for instances belonging to that class. The $K$ most frequent features were selected from the collection of instance-wise explanations and used to generate the class-wise explanation. We tested the value of $K$ in (6, 7, 8, 9, 10) for tabular datasets and in (10, 20, 30, 40, 50) for text datasets. We refer to these different settings as $K$-LIME. Class-wise explanations were produced for MUSE by grouping decision rules. Those rules leading to the same class were extracted and used as the class-wise explanation of that class. Every method approximates the behavior of the black-boxes by assigning class labels to instances based on class-wise explanations generated by the method. Table 3 presents the fidelity scores obtained by the explanation methods in class-wise experiments on the tabular and text classification datasets. For brevity reasons, when fidelity results are reported for each explanation method, the average of fidelity scores on the three black-box classifiers (MLP, SVM, and GBDT) is reported for each tabular dataset.

As the results show, our CIE method obtained the highest scores on two of the tabular datasets and on all the three text datasets. The optimization procedure utilized by CIE can efficiently select an optimal subset of instance-wise explanations that accurately approximate the decision boundaries of a specific class. Although class-wise explanations are not as accurate as instance-wise ones, they can be still useful in discriminating different classes based on important feature values and feature correlations that characterize each subspace of the predictive model.

**Table 3.** The fidelity scores obtained by the explanation methods for class-wise explanation experiments on the tabular and text datasets. The best score obtained on each dataset is shown in bold type.

|         | Tabular datasets | | | Text datasets | | |
|---------|-------|--------|-------|---------|-------|--------|
|         | Adult | Compas | TD    | TREC-QC | SST   | SMS-SC |
| MUSE    | 0.842 | 0.829  | 0.800 | 0.780   | 0.773 | 0.765  |
| 6-LIME  | 0.779 | 0.788  | 0.813 | -       | -     | -      |
| 7-LIME  | 0.779 | 0.786  | 0.820 | -       | -     | -      |
| 8-LIME  | 0.786 | 0.790  | 0.821 | -       | -     | -      |
| 9-LIME  | 0.795 | 0.803  | 0.831 | -       | -     | -      |
| 10-LIME | 0.807 | 0.811  | **0.837** | 0.604 | 0.599 | 0.591 |
| 20-LIME | -     | -      | -     | 0.694   | 0.708 | 0.681  |
| 30-LIME | -     | -      | -     | 0.839   | 0.816 | 0.815  |
| 40-LIME | -     | -      | -     | 0.850   | 0.823 | 0.838  |
| 50-LIME | -     | -      | -     | 0.857   | 0.841 | 0.844  |
| Greedy  | 0.747 | 0.729  | 0.736 | 0.789   | 0.773 | 0.777  |
| Random  | 0.469 | 0.442  | 0.485 | 0.507   | 0.478 | 0.452  |
| CIE     | **0.851** | **0.842** | 0.829 | **0.887** | **0.859** | **0.881** |

## 4.2. Interpretability

We conducted a set of experiments to assess the interpretability of the comparison methods. Given a parameter $K$ that specifies the maximum number of decision itemsets per class, we experimented the methods with different values of $K$ in the range [5, 50] with an interval of 5, in order to assess their ability in producing class-wise explanations that satisfy both the interpretability and descriptive accuracy criteria. A decision itemset refers to a confident itemset and a frequent itemset in CIE and MUSE, respectively. Since LIME uses coefficients of a linear model as explanations, we considered every individual feature and the corresponding coefficient as a decision itemset in favor of the respective class.

Figure 6 shows the descriptive accuracy against the maximum number of decision itemsets per class, i.e. the parameter $K$, for class-wise explanations produced by CIE, LIME, and MUSE. The results are reported for all the six datasets used in the previous experiments. All the interpretability experiments reported in this subsection were done using the predictions of a multi-layer perceptron made on the tabular datasets and the predictions of a LSTM made on the text datasets.

As the results show, our CIE method can achieve higher levels of descriptive accuracy than the other methods when they use the same number of decision itemsets, except for one of the tabular datasets. The results demonstrate that the explanations generated by CIE are more interpretable than those of the other comparison methods since an explanation model with fewer decision elements is considered more interpretable (Lakkaraju et al., 2017).

We also created a global explanation for every method through aggregating the class-wise explanations produced by the best setting of each method based on the results reported in Table 3. We computed the two interpretability measures, i.e. the size of explanation and the total number of items, defined by Eq. (6) and Eq. (7), for the global explanations. For brevity reasons, the results are reported for two tabular and two text datasets. Figure 7 shows the descriptive accuracy against the size of the global explanations. In order to select $K$ decision itemsets from the collection of class-wise explanations produced by a method, a score was assigned to each decision itemset based on two criteria, i.e. 1) the number of records that the itemset covers and 2) the percentage of records that are only covered by the itemset, but not by any other itemsets. The decision itemsets were ranked and the top $K$ itemsets were selected as the global explanation for the respective dataset. Figure 8 shows the descriptive accuracy against the total number of items in global explanations produced by the explanation methods.

These results demonstrate that our CIE method can produce global explanations that are more interpretable than those of the other comparison methods. As can be observed in Figure 7 and Figure 8, the difference between CIE and the other methods is higher on the text datasets than the tabular datasets. For example, CIE obtained a descriptive accuracy of 76% on TREC question classification dataset when it used 40 decision itemsets, whereas LIME and MUSE reached that descriptive accuracy score using 70 and 80 decision itemsets, respectively. The main reason for the higher performance achieved by CIE is that it utilizes confident itemsets that can effectively capture decision boundaries of the classes, and the optimization procedure that selects the minimal subset of confident itemsets that satisfy fidelity, interpretability, and coverage criteria.

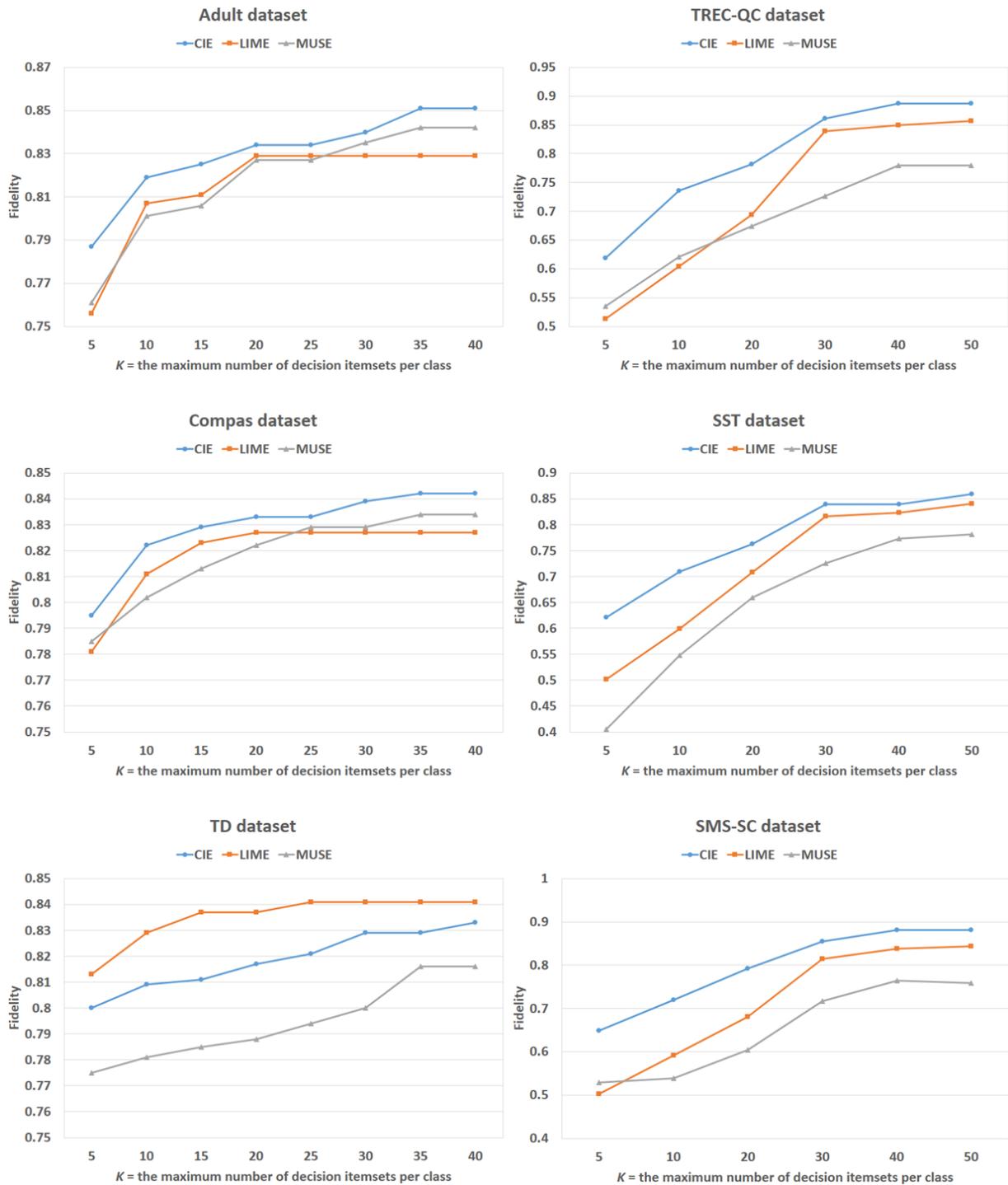

**Figure 6.** The interpretability results for class-wise explanations produced by the methods on the three tabular and the three text classification datasets. The maximum number of decision itemsets per class, i.e. the parameter *K*, is reported against the descriptive accuracy for each dataset.

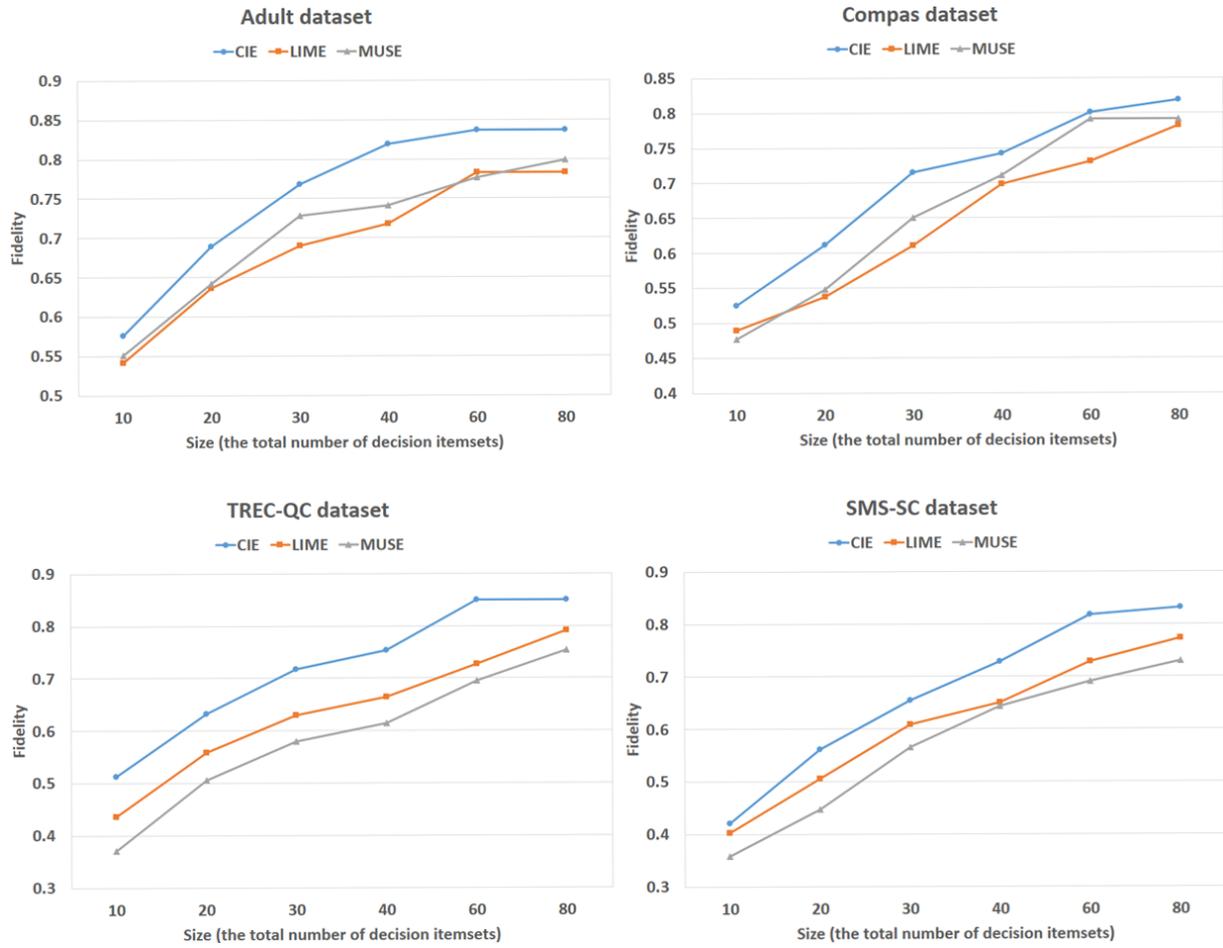

**Figure 7.** The interpretability results for class-wise explanations produced by the methods on the two tabular and the two text classification datasets. The total number of decision itemsets in a global explanation, i.e. the size of explanation, is reported against the descriptive accuracy for each dataset.

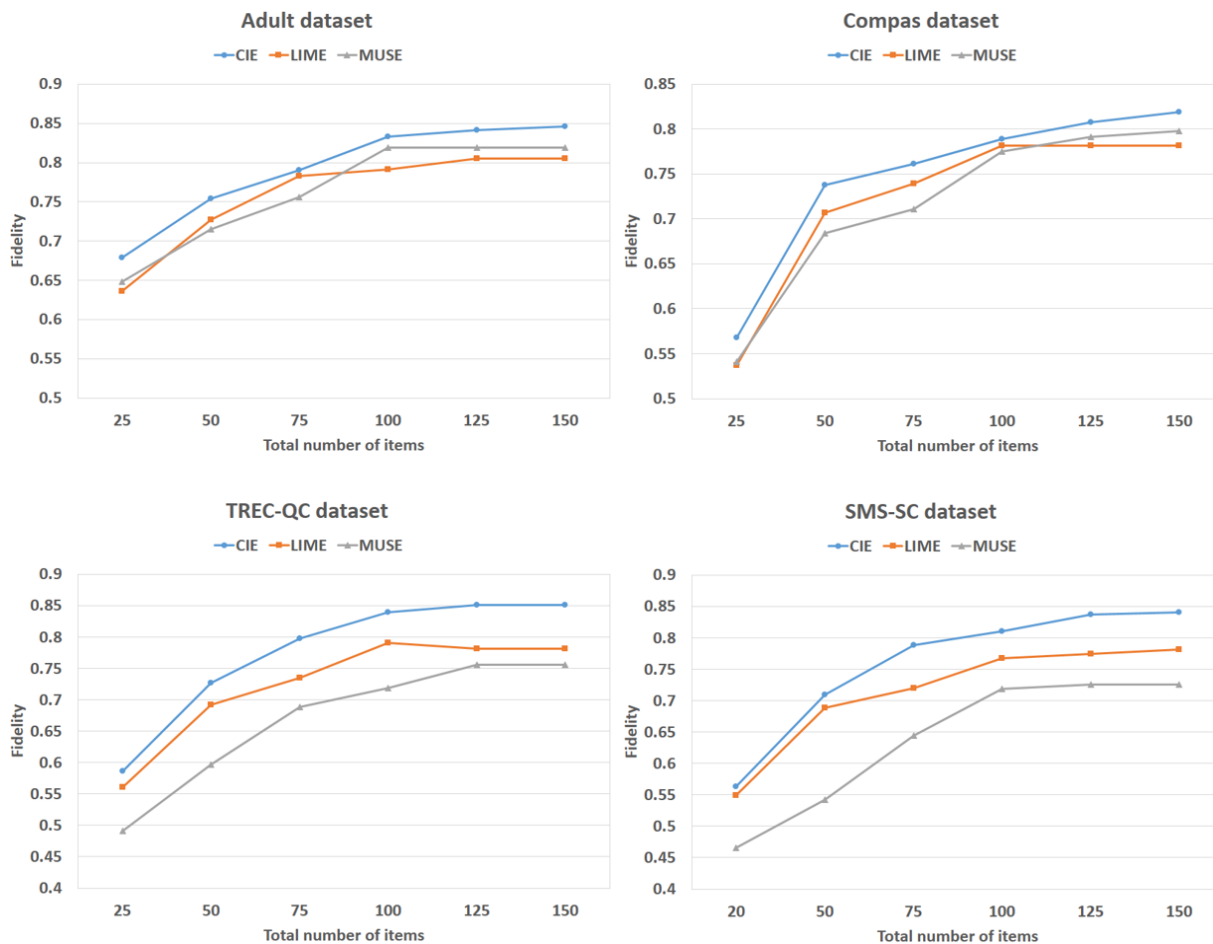

**Figure 8.** The interpretability results for class-wise explanations produced by the methods on the two tabular and the two text classification datasets. The total number of items in a global explanation is reported against the descriptive accuracy for each dataset.

## 4.3. User study

We conducted a user study with 20 participants who were researchers or postgraduate students in AI, ML or related fields. Every user was provided with 18 instances randomly chosen from the Adult, Compas, TREC-QC, and SMS-SC datasets. For every instance, the users were also given the outcome of a MLP (for the tabular datasets) or a LSTM (for the text datasets) classifier, the real class, and an explanation produced by CIE, LIME, or MUSE.

The users were asked to examine each instance and the respective explanation, then judge about the interpretability of the explanation methods. Every user assigned a score between 1 and 5 to every explanation according to four criteria: 1) how easily the user can understand the explanation, 2) how easily the user can infer about the black-box's behavior or logic, based on the explanation, 3) the size of explanation, and 4) how relevant the explanation is to the task at hand. An overall interpretability score was computed for each explanation method through averaging the scores assigned by the users to the explanations produced by that method. Table 4 presents the average interpretability scores assigned by the users. For every explanation method, an average score is separately presented for all the samples, the samples from the textual datasets, and the samples from the tabular datasets.

**Table 4.** The interpretability scores assigned by the users to the explanations produced by CIE, LIME, and MUSE (higher values are better). The best score in each column is shown in bold type.

|  | Average user scores | | |
|---|---|---|---|
|  | All samples | Textual samples | Tabular samples |
| CIE | **3.41** | **3.33** | **3.49** |
| LIME | 2.83 | 2.74 | 2.91 |
| MUSE | 1.74 | 1.24 | 2.24 |

The users assigned the highest scores to the explanations generated by CIE. According to the comments made by the users, CIE produces more interpretable explanations because the confidence score provides a straightforward measure to assess the strength of relationships between the feature values and the outcome, confident itemsets show the collective importance of multiple features on the outcome, the size of confident itemsets is smaller than the size of coefficients of a linear model or decision sets made of multiple if-then rules, and it is easier to infer about the behavior of the black-boxes by observing the explanations generated by CIE.

As can be seen in Table 4, CIE obtained the highest user scores for explaining both types of textual and tabular data records. The users assigned the lowest scores to the explanations produced by MUSE for textual data records. This shows frequent itemsets cannot be effectively used to explain text classification outcomes in an interpretable form. For all the explanators, the explanations of tabular data records obtained higher scores than those of textual samples. This shows that explaining the outcome of the text classifiers in an understandable and interpretable way is harder than explaining the outcome of the classifiers made on the tabular datasets.

## 5. Conclusion

In this paper, we have proposed CIE, a post-hoc and model-agnostic method for explaining predictions made by black-box classifiers. CIE discretized the input space by extracting confident itemsets representing those feature values that are highly associated with a class label. The extracted confident itemsets were easily used to produce concise instance-wise explanations. Class-wise explanations were produced through optimizing fidelity, interpretability, and coverage objectives in order to select minimal subsets of confident itemsets extracted for a given class.

We demonstrated the flexibility of our CIE method by explaining outcomes of various black-box classifiers on a variety of text and tabular datasets. The results of fidelity experiments showed that CIE can mimic the black-box's behavior more accurately than the other XAI methods in terms of instance-wise and class-wise explanations. The confident itemset mining utilized by CIE can efficiently approximate decision boundaries of the black-box models. This helps the explanation method identifies those feature values that are highly correlated to specific class labels assigned by the target classifier.

Comparing the number of decision itemsets and items against the descriptive accuracy of global explanations, it was shown that our CIE method can produce more interpretable explanations since it achieved higher descriptive accuracy scores than the other explanators when it used fewer decision itemsets and items. The results of user experiments showed that CIE can produce more interpretable and understandable explanations than the other comparison methods in terms of subjective evaluations made by the users. This shows that our confident itemset-based explanation method can effectively approximate and represent the black-box's behavior in an interpretable and understandable form. Regarding the experimental results, the following concluding remarks can be made:

- Confident itemsets provide an effective way of capturing local relationships between feature values and class labels in various subspaces of a black-box classifier's decision space.
- Optimizing the fidelity, interpretability, and coverage measures on the extracted local relationships can lead to producing high-quality class-wise explanations.
- The perturbation-based and decision sets explanation methods impose some limitations that decrease the descriptive accuracy and interpretability, especially when they produce class-wise explanations. The CIE method tackles those limitations by relying on real feature values (instead of perturbed features), using the confident measure to capture all important correlations in the subspaces, providing a measure of confidence that quantifies the strength of the correlations, revealing ambiguity and uncertainty in the decision space of the model or in the data, and optimizing the essential fidelity, interpretability, and coverage measures.

The next stage of our research will be modifying CIE to be applied in domain-specific tasks, e.g. biomedical natural language processing. Incorporating sources of domain knowledge into the explanation method will be required to build a semantical representation of the data in addition to word-based and syntactical features (Gao, Liu, Lawley, & Hu, 2017). This helps reveal various types of lexical, semantical, and syntactical relationships learned by the target black-box. One other future line of work is to augment CIE by adding a mechanism that shows what modifications in the feature values of a data record are needed to change the outcome of the black-box. Such a mechanism would be especially useful in decision support or recommender systems where the user needs to be provided with information about what data relationships should exist or not exist in order to change the class of a record, or what feature values should change to convert a true negative sample into a positive one (Tolomei et al., 2017).